\documentclass{amsart}
\usepackage{amsmath}
\usepackage{amssymb}
\usepackage{listings}
\usepackage{algorithm}
\usepackage{algorithmic}

\newtheorem{thm}{Theorem}[section]
\newtheorem{cor}[thm]{Corollary}
\newtheorem{lem}[thm]{Lemma}

\newtheorem{defn}[thm]{Definition}

\newcommand{\Real}{\mathbb R}

\renewcommand{\b}[1]{\mathbf{#1}}
\newcommand{\bx}{\b x}
\newcommand{\by}{\b y}
\newcommand{\bigtimes}{\mathop{\times}\limits}
\newcommand{\maxnorm}[1]{\left\|#1\right\|_\infty}
\newcommand{\asize}{|A|}

\newcommand{\rmax}{R_{\max}}
\newcommand{\vmax}{V_{\max}}

\newcommand{\ro}{R_\textrm{E}}

\newcommand{\citet}[1]{\cite{#1}}
\newcommand{\citep}[1]{\cite{#1}}

\newcommand{\ext}[1]{E_{[#1]}}

\newcommand{\bproof}{\textit{Proof. }}
\newcommand{\eproof}{\hfill $\blacksquare$}

\renewcommand{\hat}[1]{\widehat{#1}}

\begin{document}

%
%

\title[Optimistic Initialization for FMDPs -- Extended version]{Optimistic Initialization and Greediness Lead to Polynomial Time Learning in
Factored MDPs \\ Extended Version}%
\author{Istv{\'a}n Szita \and Andr{\'a}s L{\H{o}}rincz}
\date{}

\maketitle

\begin{abstract}
In this paper we propose an algorithm for polynomial-time reinforcement learning in factored Markov
decision processes (FMDPs). The factored optimistic initial model (FOIM) algorithm, maintains an
empirical model of the FMDP in a conventional way, and always follows a greedy policy with respect
to its model. The only trick of the algorithm is that the model is initialized optimistically. We
prove that with suitable initialization (i) FOIM converges to the fixed point of approximate value
iteration (AVI); (ii) the number of steps when the agent makes non-near-optimal decisions (with
respect to the solution of AVI) is polynomial in all relevant quantities; (iii) the per-step costs
of the algorithm are also polynomial. To our best knowledge, FOIM is the first algorithm with
these properties.
This extended version contains the rigorous proofs of the main theorem. A version of this paper
appeared in ICML'09.
\end{abstract}

\section{Introduction}

Factored Markov decision processes (FMDPs) are practical ways to compactly formulate sequential
decision problems---provided that we have ways to
solve them. When the environment is unknown, all effective reinforcement learning methods apply
some form of the ``optimism in the face of uncertainty'' principle: whenever the learning agent
faces the unknown, it should assume high rewards in order to encourage exploration. \emph{Factored
optimistic initial model} (FOIM) takes this principle to the extreme: its model is initialized to
be overly optimistic. For more often visited areas of the state space, the model gradually gets
more realistic, inspiring the agent to head for unknown regions and explore them, in search of some
imaginary ``Garden of Eden''. The working of the algorithm is simple to the extreme: it will not
make any explicit effort to balance exploration and exploitation, but always follows the greedy
optimal policy with respect to its model. We show in this paper that this simple (even simplistic) trick is sufficient for
effective FMDP learning.

The algorithm is an extension of OIM (\emph{optimistic initial model}) \citep{Szita08Many}, which
is a sample-efficient learning algorithm for flat MDPs. There is an important difference, however,
in the way the model is solved. Every time the model is updated, the corresponding value function
needs to be re-calculated (or updated) For flat MDPs, this is not a problem: various dynamic
programming-based algorithms (like value iteration) can solve the model to any required accuracy in
polynomial time.

The situation is less bright for generating near-optimal FMDP solutions: all currently known
algorithms may take exponential time, e.g. the approximate policy iteration of
\citet{Boutilier00Stochastic} using decision-tree representations of policies, or solving the
exponential-size flattened version of the FMDP. If we require polynomial running time (as we do in
this paper in search for a practical algorithm), then we have to accept
sub-optimal solutions. The only known example of a polynomial-time FMDP planner is \emph{factored
value iteration} (FVI) \cite{Szita08Factored}, which will serve as the base planner for our
learning method. This planner is guaranteed to converge, and the error of its solution is bounded
by a term depending only on the quality of function approximators.

Our analysis of the algorithm will follow the established techniques for analyzing
sample-efficient reinforcement learning (like the works of
\citet{Kearns98Near-Optimal,Brafman01R-MAX,Kakade03Sample,Strehl05Theoretical,Szita08Many} on flat
MDPs and \citet{Strehl07Model-Based} on FMDPs). However, the listed proofs of convergence rely
critically on access to a near-optimal planner, so they have to be generalized suitably. By doing
so, we are able to show that FOIM converges to a bounded-error solution in polynomial time with
high probability.

We introduce basic concepts and notations in section \ref{s:notations}, then in section
\ref{s:literature} we review existing work, with special emphasis to the immediate ancestors of our
method. In sections \ref{s:FOIM_blocks} and \ref{s:foim} we describe the blocks of FOIM and the
FOIM algorithm, respectively. We finish the paper with a short analysis and discussion.

\section{Basic concepts and notations} \label{s:notations}


An MDP is characterized by a quintuple $(\b X, A, R, P, \gamma)$, where $\b X$ is a finite set of
states; $A$ is a finite set of possible actions; $R: \b X \times A \to \Real$ is the reward
function of the agent; $P: \b X \times A \times \b X \to [0,1]$ is the transition function; and
finally, $\gamma\in [0,1)$ is the discount rate on future rewards. A (stationary, Markov) policy
of the agent is a mapping $\pi: \b X \times A \to [0,1]$. The optimal value function $V^*: \b X
\to \Real$ gives the maximum attainable total rewards for each state,
and satisfies the Bellman equation
\begin{equation} \label{e:V*_bellman}
  V^*(\bx) = \max_a \sum_{\by} P(\by \mid\bx,a) \Bigl( R(\bx,a) +
  \gamma V^*(\by) \Bigr).
\end{equation}
Given the optimal value function, it is easy to get an optimal policy:
$\pi^*(\bx, a) :=1$ iff $a = \mathrm{arg}\max_a \sum_{\by} P(\by \mid\bx,a) \Bigl( R(\bx,a) +
  \gamma V^*(\by) \Bigr)$ and $0$ otherwise.
%

\subsection{Vector notation}

Let $N:=|\b X|$, and suppose that states are integers from 1 to $N$, i.e. $\b X =
\{1,2,\ldots,N\}$. Clearly, value functions are equivalent to $N$-dimensional vectors of reals,
which may be indexed with states. The vector corresponding to $V$ will be denoted as $\b v$ and
the value of state $\bx$ by ${\b v}_{\bx}$. Similarly, for each $a$ let us define the
$N$-dimensional column vector $\b r^a$ with entries $\b r^a_\bx = R(\bx,a)$ and $N\times N$ matrix
$P^a$ with entries $P^a_{\bx,\by} = P(\by \mid \bx,a)$.

The Bellman equations can be expressed in vector notation as
\(
  \b v^* = \textbf{max}_{a\in A}  \bigl( \b r^a + \gamma P ^a
   \b v^* \bigr),
\)
where $\textbf{max}$ denotes the componentwise maximum operator. The Bellman equations are the
basis to many RL algorithms, most notably, value iteration:
\begin{equation}
  \b v_{t+1} := \textbf{max}_{a\in A}  \bigl( \b r^a + \gamma P ^a
   \b v_t \bigr),
\end{equation}
which converges to $\b v^*$ for any initial vector $\b v_0$.


\subsection{Factored structure}

We assume that $\b X$ is the Cartesian product of $m$ smaller state spaces (corresponding to
individual variables):
\[
  \b X = X_1 \times X_2 \times \ldots \times X_m.
\]
For the sake of notational convenience we will assume that each $X_i$ has the same size, $|X_1| =
|X_2| = \ldots = |X_m| = n$. With this notation, the size of the full state space is $N = |\b X| =
n^m$. We note that all derivations and proofs carry through to different size variable spaces.

\begin{defn}
For any subset of variable indices $Z \subseteq \{ 1,2,\ldots,m\}$, let $\b X[Z] :=
\bigtimes_{i\in Z} X_i $, furthermore, for any $\bx \in \b X$, let $\bx[Z]$ denote the value of
the variables with indices in $Z$. We shall also use the notation $\bx[Z]$ without specifying a
full vector of values $\bx$, in such cases $\bx[Z]$ denotes an element in $\b X[Z]$. For
single-element sets $Z=\{i\}$ we shall also use the shorthand $\bx[\{i\}] = \bx[i]$.
\end{defn}

\begin{defn}[Local-scope function]
A function $f$ is a \emph{local-scope} function if it is defined over a subspace $\b X[Z]$ of the
state space, where $Z$ is a (presumably small) index set.
\end{defn}
If $|Z|$ is small, local-scope functions can be represented efficiently, as they can take only
$n^{|Z|}$ different values.

\begin{defn}[Extension]
For $f: \b X[Z] \to \Real$ be a local-scope function. Its extension to the whole state space is
defined by $f(\bx) := f(\bx[Z])$.
The extension operator for $Z$ is a linear operator with a matrix $\ext{Z} \in \Real^{|\b
X|\times|\b X[Z]|}$, with entries
\[
  \left(\ext{Z}\right)_{\b u, \b v[Z]} = \left\{%
\begin{array}{ll}
    1, & \hbox{if $\b u[Z] = \b v[Z]$;} \\
    0, & \hbox{otherwise.} \\
\end{array}%
\right.
\]

For any local-scope function $f$ with a corresponding vector representation $\b f \in \Real^{|\b
X[Z]| \times 1}$, $\ext{Z} \b f\in \Real^{|\b X| \times 1}$ is the vector representation of the
extended function.
\end{defn}

We assume that the reward function is the sum of $J$ local-scope functions with scopes $Z_j$: \(
  R(\bx, a) = \sum_{j=1}^{J} R_j(\bx[Z_j], a).
\) In vector notation:  $\b r^a = \sum_{j=1}^J \ext{Z_j^a} \b r^a_i$.
We also assume that for each variable $i$ there exist neighborhood sets $\Gamma_i$ such that the value
of $\bx_{t+1}[i]$ depends only on $\bx_{t}[\Gamma_i]$ and the action $a_t$ taken. Then we can
write the transition probabilities in a factored form
\begin{equation} \label{e:Pfactored}
  P(\by \mid \bx, a) = \prod_{i=1}^m P_i(\by[i] \mid \bx[\Gamma_i], a)
\end{equation}
for each $\bx,\by\in\b X$, $a\in A$, where each factor is a local-scope function $P_i : \b
X[\Gamma_i] \times A \times X_i \to [0,1]$ (for all \mbox{$i \in \{1,\ldots,m\}$}). In
vector/matrix notation, for any vector $\b v \in \Real^{|\b X|\times 1}$,
 \(
  P^a \b v = \bigotimes_{i=1}^{m} (P^a_i \b v[\Gamma_i]),
 \)
where $\bigotimes$ denotes the Kronecker product.
Finally, we assume that the size of all local scopes are bounded by a small constant $m_f \ll m$:
$|\Gamma_i| \leq m_f$ for all $i$. As a consequence, all probability factors can be represented
with tables having at most $N_f := n^{m_f}$ rows.

An FMDP is fully characterized by the tuple $\mathcal M = \bigl( \{X_i\}_{i=1}^m; A;
\{Z_j\}_{j=1}^{J}; \{R_j\}_{j=1}^{J}; \{\Gamma_i\}_{i=1}^m; \{P_i\}_{i=1}^m; \bx_s;
     \gamma\bigr)$.

\section{Related literature} \label{s:literature}

The idea of representing a large MDP using a factored model was first proposed by
\citet{Koller00Policy} but similar ideas appear already in the works of
\citet{Boutilier95Exploiting,Boutilier00Stochastic}.

\subsection{Planning in known FMDPs}


Decision trees (or equivalently, decision lists) provide a way to represent the agent's policy
compactly. \citet{Koller00Policy} and \citet{Boutilier95Exploiting,Boutilier00Stochastic} present
algorithms to evaluate and improve such policies, according to the policy iteration scheme.
Unfortunately, the size of the policies may grow exponentially even with a decision tree
representation \citep{Boutilier00Stochastic,Liberatore02Size}.

The exact Bellman equations (\ref{e:V*_bellman}) can be transformed to an equivalent linear program
with $N$ variables and $N\cdot |A|$ constraints. In the approximate linear programming approach, we
approximate the value function as a linear combination of $K$ basis functions, resulting in an
approximate LP with $K$ variables and $N\cdot |A|$ constraints. Both the objective function and the
constraints can be written in compact forms, exploiting the local-scope property of the appearing
functions. \citet{Guestrin02Efficient} show that the maximum of exponentially many local-scope
functions can be computed by rephrasing the task as a non-serial dynamic programming task and
eliminating variables one by one. Therefore, the equations can be transformed to an equivalent,
more compact linear program. The gain may be exponential, but this is not necessarily so in all
cases. Furthermore, solutions will not be \mbox{(near-)optimal} because of the function
approximation; the best that can be proved is bounded error from the optimum (where the bound
depends on the quality of basis functions used for approximation).

The approximate policy iteration algorithm \citep{Koller00Policy,Guestrin02Efficient} also uses an
approximate LP reformulation, but it is based on the policy-evaluation Bellman equations.
Policy-evaluation equations are, however, linear and do not contain the maximum operator, so there
is no need for a costly transformation step. On the other hand, the algorithm needs an explicit
decision tree representation of the policy. \citet{Liberatore02Size} has shown that the size of
the decision tree representation can grow exponentially. Furthermore, the convergence properties
of these algorithms are unknown.

Factored value iteration \cite{Szita08Factored} also approximates the value function as a linear
combination of basis functions, but uses a variant of approximate value iteration:
the projection operator is modified to avoid divergence. FVI converges in a polynomial
number of steps, but the solution may be sub-optimal. The error of the solution has bounded
distance from the optimal value function, where the bound depends on the quality of function
approximation. As an integral part of FOIM, FVI is described in detail in Section~\ref{ss:fvi}.

\subsection{Reinforcement Learning in FMDPs}

In the reinforcement learning setting, the agent interacts with an FMDP environment with unknown
parameters. In the model-based approach, the agent has to learn the structure of the FMDP (i.e.,
the dependency sets $\Gamma_i$ and the reward domains $Z_j$), the transition probability factors
$P_i$ and the reward factors $R_j$.

\textbf{Unknown transitions.} Most approaches assume that the structure of the FMDP and the reward
functions are known, so only transition probabilities need to be learnt. Examples include the
factored versions of sample-efficient model-based RL algorithms: factored E$^3$
\citep{Kearns99Efficient}, factored R-max \citep{Guestrin02Algorithm-Directed}, or factored MBIE
\citep{Strehl07Model-Based}. All the abovementioned algorithms have polynomial
sample complexity (in all relevant task parameters), and require polynomially many calls to an
FMDP-planner. Note however, that all of the mentioned approaches require access to a planner that
is able to produce $\epsilon$-optimal solutions\footnote{The assumption of
\citep{Kearns99Efficient} is slightly less restrictive: they only require that the value of the
returned policy has value at least $\rho V^*$ with some $\rho<1$. However, no planner is known that
can achieve this and cannot achieve near-optimality.} -- and to date, no algorithm exists that
would accomplish this accuracy in polynomial time. \cite{Guestrin02Algorithm-Directed} also present
an algorithm where exploration is guided by the uncertainties of the linear programming solution.
While this approach does not require access to a near-optimal planner, no formal performance bounds
are known.

\textbf{Unknown rewards.} Typically, it is asserted that the rewards can be approximated from
observations analogously to transition probabilities. However, if the reward is composed of
multiple factors (i.e., $J>1$), then we can only observe the
\emph{sums} of unknown quantities, not the individual quantities themselves. To date, we know of no
efficient approximation method for learning factored rewards.

\textbf{Unknown structure.} Few attempts exist that try to obtain the structure of the FMDP
automatically. \citet{Strehl07Efficient} present a method that learns the structure of an FMDP in
polynomial time (in all relevant parameters).

\section{Building blocks of FOIM}\label{s:FOIM_blocks}

We describe the two main building blocks of our algorithm, \emph{factored value iteration} and \emph{optimistic initial model}.

\subsection{Factored value iteration} \label{ss:fvi}


We assume that all value functions are approximated as the linear combination of $K$ basis functions $h_k: \b X
\to \Real$: $V(\bx) = \sum_{k=1}^K w_k h_k(\bx)$.

Let $H$ be the $N\times K$ matrix mapping feature weights to state values, with entries $H_{\bx,k}
= h_k(\bx)$, and let $G$ be an arbitrary $K\times N$ linear mapping projecting state values to
feature weights. Let $\b w \in \Real^{K}$ denote the weight vector of the basis functions. It is
known that if $\maxnorm{HG} \leq 1$, then the approximate Bellman equations
 $
  \b w^\times = G \textbf{max}_{a\in A}  \bigl( \b r^a + \gamma P ^a
   H \b w^\times \bigr)
 $
have a unique fixed point solution $\b w^\times$, and \emph{approximate value iteration (AVI)}
\begin{equation} \label{e:avi_vectorized}
  \b w_{t+1} := G \textbf{max}_{a\in A}  \bigl( \b r^a + \gamma P ^a
   H \b w_t \bigr)
\end{equation}
converges there for any starting vector $\b w_0$.

\begin{defn}
Let the AVI-optimal value function be defined as $\b v^\times = H \b w^\times$.
\end{defn}
As shown by \citet{Szita08Factored}, the distance of AVI-optimal value function from the true
optimum is bounded by the projection error of $\b v^*$:
\begin{equation} \label{e:vtimes_errorbound}
  \maxnorm{\b v^\times - \b v^*} \leq {\textstyle \frac{1}{1-\gamma}} \maxnorm{H G \b
  v^* - \b v^*}.
\end{equation}


We make the further assumption that all the basis functions are local-scope ones: for each
$k\in \{1,\ldots,K\}$, $h_k: \b X[C_k] \to \Real$, with feature matrices $H_k \in \Real^{|\b
X[C_k]| \times K}$. The feature matrix $H$ can be decomposed as $H = \sum_{k=1}^K \ext{C_k} H_k$.

\begin{defn}
For any matrices $H$ and $G$, let the row-normalization of $G$ be a matrix $\mathcal{N}(G)$
of the same size as $G$, and having the entries \(
  [\mathcal{N}(G)]_{k,\bx} = \frac{G_{k,\bx}}{\maxnorm{[HG]_{k,*}}}.
\)
\end{defn}
Throughout the paper, we shall use the projection matrix $G = \mathcal{N}( H^T)$.

The AVI equation (\ref{e:avi_vectorized}) can be considered as the product of the $K\times N$
matrix $G$ and an $N\times 1$ vector $\b v_t = \textbf{max}_{a\in A}  \bigl( \b r^a + \gamma P ^a
H \b w_t \bigr)$. Using the above assumptions and notations, we can see that for any $\bx \in \b
X$, the corresponding columm of $G$ and the corresponding element of $\b v_t$ can be computed in
polynomial time:
\begin{align*}
 &[G]_{k,\bx} =
 \frac{1}{\maxnorm{[HH^T]_{k,\ast}}}\sum_{k'=1}^K \left[H_{k'}^T\right]_{\ast,\bx[C_{k'}]}; \\
 &{}[\b v_t]_{\bx}
        = \max_{a\in A} \Bigl[ \sum_{j=1}^J [\b r^a_j]_{\bx[Z_j^a]}\! + \!\gamma\! \sum_{k=1}^K \ext{\Gamma_{\cup C_k}
} \bigl( \!\bigotimes_{i\in C_k}\! P^a_i \bigr)  (\b h_k w_{k,t})
    \Bigr]
\end{align*}


Factored value iteration draws $N_1 \ll N$ states uniformly at random, and performs approximate
value iteration on this reduced state set.

\begin{thm}[\cite{Szita08Factored}]
Suppose that $G = \mathcal{N}(H^T)$ For any $\epsilon>0$, $\delta>0$, if the sample size is $N_1 =
O(\frac{m^2}{\epsilon^2} \log \frac{m}{\delta})$, then with probability at least $1-\delta$,
factored value iteration converges to a weight vector $\b w$ such that $\maxnorm{\b w-\b
w^\times}\leq \epsilon$. In terms of the optimal value function,
\begin{equation}
  \maxnorm{\b v^\times - \b v^*} \leq {\textstyle\frac{1}{1-\gamma}} \maxnorm{H G \b
  v^* - \b v^*} + \epsilon.
\end{equation}
\end{thm}

\subsection{Optimistic initial model for flat MDPs}

There are a number of sample-efficient learning algorithms for MDPs, e.g., E3, Rmax, MBIE, and
most recently, OIM. The underlying principle of all these methods is similar: they all maintain an
approximate MDP model of the environment. Wherever the uncertainty of the model parameters is
high, the models are optimistic. This way, the agent is encouraged to explore the unknown areas,
reducing the uncertainty of the models.

Here, we shall use and extend OIM to factored environments. In the
OIM algorithm, we introduce a hypothetical ``garden of Eden'' (GOE) state $x_E$, where the agent
gets a very large reward $\ro$ and remains there indefinitely. The model is initialized with fake
experience, according to which the agent has experienced an $(x,a,x_E)$ transition for all $x\in X$
and $a\in A$. According to this initial model, each state has value $\ro/(1-\gamma)$, which is a
major overestimation of the true values. The model is continuously updated by the collected
experience of the agent, who always takes the greedy optimal action with respect to its current
model. For well-explored $(x,a)$ pairs, the optimism of the model vanishes, thus encouraging the
agent to explore the less-known areas.

The reason for choosing OIM is twofold: (1) The optimism of the model is ensured at initialization
time, and after that, no extra work is needed to ensure the optimism of the model or to encourage
exploration. (2) Results on several
standard benchmark MDPs indicate that OIM is superior to the other algorithms
mentioned.

\section{Learning in FMDPs with an Optimistic initial model} \label{s:foim}

Similarly to other approaches, we will make the assumptions that (a) the dependencies are known,
and (b) the reward function is known, only the transition probabilities need to be learned.

\subsection{Optimistic initial model for factored MDPs}

During the learning process, we will maintain approximations of the model, in particular, of the
transition probability factors. We extend all state factors with the hypothetical "garden of Eden"
state $x_E$. Seeing the current state $\bx$ and the action $a$ taken, the transition model should
give the probabilities of various next states $\by$. Specifically, the $i$th factor of the
transition model should give the probabilities of various $y_i$ values, given $\bx[\Gamma_i]$ and
$a$. Initially, the agent has no idea, so we let it start with an overly optimistic model: we
inject the fake experience to the model that taking action $a$ in
$\bx[\Gamma_i]$ leads to a state with $i$th component $y_i = x_E$. This optimistic model will
encourage the agent to explore action $a$ whenever its  state
is consistent with $\bx[\Gamma_i]$. After many visits to
$(\bx[\Gamma_i],a)$, the weight of the initial fake experience will shrink, and the optimistic
belief of the agent (together with its exploration-boosting effect) fades away. However, by that
time, the collected experience provides an accurate approximation of the $P_i( y_i \mid
\bx[\Gamma_i],a)$ values.

So, according to the initial model (based purely on fake experience),
\[
  \hat{P}(\b y \mid \bx, a) =\left\{%
\begin{array}{ll}
    1, & \hbox{if $\b y = (x_E, \ldots, x_E)$;} \\
    0, & \hbox{otherwise,} \\
\end{array}%
\right.
\]
$
   \hat{R}(\bx, a) =
    c\cdot\ro,
$
if $c$ components of $\bx$ are $x_E$. This model is optimistic indeed, all non-GOE states have value at least $\gamma\ro/(1-\gamma) $.
Note that it is not possible to encode the $\ro$-rewards for the $GOE$ states using the original
set of reward factors, so for all state factor $i$, we add a new reward factor with local scope
$X_i$: $R'_i : X_i \times A \to \Real$, defining
\(
   R'_i(x, a) =\left\{%
\begin{array}{ll}
    \ro, & \hbox{if $x = x_E$;} \\
    0, & \hbox{otherwise.} \\
\end{array}%
\right.
\)
With this modification, we are able to fully specify our algorithm, as shown in the pseudocode
below.

\begin{algorithm}[h!]
\caption{Factored optimistic initial model.} \label{alg:foim}
\begin{algorithmic}
     \STATE \textbf{input:}
     \STATE $\mathcal M = \bigl( \{X_i\}_{1}^m; A; \{Z_j\}_{1}^{J}; \{R_j\}_{1}^{J};
\{\Gamma_i\}_{1}^m; \{P_i\}_{1}^m; \bx_s;
     \gamma\bigr)$
     \STATE $\{H_k\}_{1}^{K}$; $\{C_k\}_{1}^K$; $\epsilon>0$;
$\delta>0$; $\ro$
     \STATE \textbf{initialization:}
    \STATE $t := 0$;

    \STATE for all $i$, add GOE states: $X_i :=  X_i \cup \{x_E\}$

    \STATE for all $i$, add GOE reward function $\b r'_i$

    \item for all $i$, $a$, $\bx[\Gamma_i]$, $y\in X_i\setminus
    \{x_E \}$, let
    \STATE $\textit{TransitionCount}(\bx[\Gamma_i],a,y) := 0;$
    \STATE $\textit{TransitionCount}(\bx[\Gamma_i],a,x_E) := 1;$
    \STATE $\textit{VisitCount}(\bx[\Gamma_i],a) := 1;$

     \REPEAT{}
         \STATE $[\hat P^a_i]_{\bx[\Gamma_i],y} :=
         \frac{\textit{TransitionCount}_i(\bx[\Gamma_i],a,y)}{\textit{VisitCount}_i(\bx[\Gamma_i],a)}$.

         \STATE $\b w_t := \textit{FactoredValueIteration}(\hat{\mathcal{M}}, \{\hat P_i^a\}, \epsilon, \delta)$
         \STATE update $\textit{TransitionCount}$ and $\textit{VisitCount}$
        corresponding to transition $(\bx_t, a_t, \bx_{t+1})$.
         \STATE $t := t+1$
     \UNTIL{interaction lasts}
\end{algorithmic}
\end{algorithm}

\subsection{Analysis}

Below we prove that FOIM gets as good as possible.
What is ``as good as possible''? We clearly cannot expect better policies than the one the planner
would output, were the parameters of the FMDP known. And because of the polynomial-running-time
constraint on the planner, it will not be able to compute a near-optimal solution. However, we can
prove that FOIM gets $\epsilon$-close to the solution of the planner (which is AVI-near-optimal if
the planner is FVI), except for a polynomial number of mistakes during its run.\footnote{We are using the term
\emph{polynomial} and \emph{polynomial in all relevant quantities} as a shorthand for
\emph{polynomial in $m$, $N_f$, $\asize$, $\rmax$, $1/(1-\gamma)$, $1/\epsilon$ and $1/\delta$}.}

\begin{thm}
Suppose that an agent is following FOIM in an unknown FMDP, where all reward components fall into
the interval $[0,\rmax]$, there are $m$ state factors, and all probability- and reward-factors
depend on at most $m_f$ factors. Let $N_f = n^{m_f}$ and let $\epsilon>0$ and $\delta>0$. If the
initial values of FOIM satisfy
\[ \textstyle
R_E = c\cdot \frac{m\rmax^2}{(1-\gamma)^4\epsilon} \left[\log \frac{mN_f\asize}{(1-\gamma)\epsilon\delta}\right]
\,,
\]
then the number of timesteps when FOIM makes non-AVI-near-optimal moves, i.e., when
$
  Q^\textit{FOIM}(\bx_t,a_t) < Q^\times(\bx_t,a_t) - \epsilon \,,
$
is bounded by
\[\textstyle
  O\left( \frac{   \rmax^2 m^4 N_f\asize   }{\epsilon^4(1-\gamma)^4} \log^3\frac{1}{\delta} \log^2 \frac{mN_f\asize}{\epsilon}
  \right)
\]
with probability at least $1-\delta$.
\end{thm}

\emph{Proof sketch.} The proof uses standard techniques from the literature of sample-efficient
reinforcement learning. Most notably, our proof follows the structure of
\citet{Strehl07Model-Based}. There are two important differences compared to previous approaches:
(1) we may not assume that the planner is able to output a near-optimal solution, and (2) FOIM may
make an unbounded number of model updates, so we cannot make use of the standard argument that ``we
are encountering only finitely many different models, each of them fails with negligible
probability, so the whole algorithm fails with negligible probability''. Instead, a more careful
analysis of the failure probability is needed.  The rigorous proof can be found in the appendix.

\subsubsection{Boundedness of value functions}

According to our assumptions, all rewards fall between $0$ and $\rmax$.
From
this, it is easy to derive an upper bound on the magnitude of the AVI-optimal value function $\b
v^\times$. The bound we get is $\maxnorm{\b v^\times} \leq \frac{3-\gamma}{1-\gamma} \vmax :=
V_0$. For future reference, we note that $V_0 = \Theta(\frac{\rmax}{(1-\gamma)^2})$.


\subsubsection{From visit counts to model accuracy}

The FOIM algorithm builds a transition probability model by keeping track of visit counts to
state-action components $(\bx[\Gamma_i],a)$ and state-action-state transition components
$(\bx[\Gamma_i],a,y)$. First of all, we show that if a state-action component is visited many
times, then the corresponding probability components $\hat P_{t,i}(y|\bx[\Gamma_i],a)$ become
accurate.

Let us fix a timestep $t\in\mathbb N$, a probability factor $i \in \{1,\ldots,m\}$ and a
state-action component $(\bx[\Gamma_i],a)  \in \b X[\Gamma_i]\times A$, and $\epsilon_t>0$. Let us
denote the number of visits to the component up to time $t$ by $k_t(\bx[\Gamma_i],a)$. Let us
introduce the shorthands $p_{i} = P_{i}(y| \bx[\Gamma_i], a)$ and
$\hat p_{t,i} = \hat P_{t,i}(y| \bx[\Gamma_i], a)$. By
Theorem 3 of \citet{Strehl07Model-Based} (an application of the Hoeffding--Azuma inequality),
\begin{equation} \label{es:strehl_thm3}
  \Pr\Bigl( \sum_{y \in X_i} \left| p_i( - \hat p_{t,i} \right| \!>\! \epsilon_t \Bigr)
  \leq 2^n {\textstyle\exp\Bigl(-\frac{\epsilon_t^2 k_t(\bx[\Gamma_i],a) }{2}\Bigr)} \,.
\end{equation}
Unfortunately, the above inequality only speaks about a single time step $t$, but we need to
estimate the failure probability for the whole run of the algorithm.
By the union bound, that is at most
\begin{eqnarray} \label{es:epsilondelta_probsum}
   \sum_{k=1}^\infty \Pr\Bigl( \sum_{y \in X_i} \left| p_i - \hat p_{t_k,i} \right| >
   \epsilon_{t_k}  \Bigr) \,.
\end{eqnarray}
Let $k_0 := \Theta\left(\frac{m^2}{(1-\gamma)^2\epsilon^2}  \log \frac{m^2N_f\asize}{(1-\gamma)\delta\epsilon}\right)$.
For $k<k_0$, the number of visits is too low, so in eq.~(\ref{es:strehl_thm3}), either $\epsilon_1$, or the right-hand side
is too big. We choose the former: we make the failure probability less than some constant
$\delta'$ by setting $\epsilon_{t_k} =
\frac{\beta(\delta')}{\sqrt{k}}$, where $\beta(\delta') = \sqrt{2(\log \frac{1}{\delta'}+ n\log
2)}$.
For $k\geq k_0$, the number of visits is sufficiently large, so we can decrease either the
accuracy or the failure probability (or even both). It turns out that an approximation accuracy
$\epsilon_{t_k} = \epsilon(1-\gamma)/m$ is sufficient, so we decrease failure probability.
Let us set $\delta' := \Theta \left( \frac{\delta\epsilon^2(1-\gamma)^2}{m^3N_f\asize}/{\log
\frac{m^2N_f\asize}{(1-\gamma)\delta\epsilon}} \right)$. With this choice of $\delta'$ and $k_0$,
$\beta(\delta')\leq\epsilon(1-\gamma)/m$ whenever $k\geq k_0$, furthermore, $2^n
\exp\left(-\frac{k\epsilon^2}{2m^2}\right) \leq \delta'$, so
we get that
\begin{align*}
  & \sum_{k=1}^\infty \Pr\Bigl( \sum_{y \in X_i} \left| p_i - \hat p_{t_k,i} \right| >
   {\textstyle\max(\frac{\beta(\delta')}{\sqrt{k}}, \frac{\epsilon(1-\gamma)}{m})}  \Bigr) \\
  & \leq \sum_{k=1}^{k_0-1} \delta' + \sum_{k=k_0}^\infty 2^n
  {\textstyle \exp\left(-\frac{k\epsilon^2}{2m^2}\right)} \\
  &
  \leq \delta' \Bigl({\textstyle k_0 +  \frac{1}{1-\exp\left(-\frac{\epsilon^2}{2m^2}\right)}}\Bigr)
  \leq
{\textstyle \Theta\left(\frac{\delta}{mN_f\asize}\right)}
  \,.
\end{align*}
We can repeat this estimation for every state-action components $(\bx[\Gamma_i],a)$. There are at
most $mN_f\asize$ of these, so the total failure probability is still less than $\Theta(\delta)$.
This means that
\begin{equation} \label{es:Pi_bound}
 \sum_{y \in X_i} \left| p_i - \hat p_{t,i} \right|
 \leq
   \max(\frac{\beta(\delta')}{\sqrt{k_t(\bx[\Gamma_i],a)}}, \frac{\epsilon(1-\gamma)}{m})
\end{equation}
will hold for \emph{all} $(\bx[\Gamma_i],a)$ pairs and \emph{all} timesteps $t$ with high
probability. From now on, we will consider only realizations where the failure event does not
happen, but bear in mind that all our statements that are based on (\ref{es:Pi_bound}) are true
only with $1-\Theta(\delta)$ probability.

From (\ref{es:Pi_bound}), we can easily get $L_1$ bounds on the accuracy of the full transition
probability function:
$
 \sum_{\by \in \b X} \left| P(\by|\bx, a) - \hat P_{t}(\by| \bx, a) \right|
 \leq $ $
   \sum_{i=1}^m \max(\frac{\beta(\delta')}{\sqrt{k_t(\bx[\Gamma_i],a)}}, \frac{\epsilon(1-\gamma)}{m}) \,
$ for all $(\bx,a) \in \b X\times A$ and for all $t$.


\subsubsection{The known-state FMDP}

A state-action component $(\bx[\Gamma_i],a)$ is called \emph{known} at timestep $t$ if it has been
visited at least $k_0$ times, i.e., if $k_t(\bx[\Gamma_i],a) \leq k_0$. We define the
\emph{known-component} FMDP $M^{K_t}$ as follows: (1) its state and action space, rewards, and the
decompositions of the transition probabilities (i.e., the dependency sets $\Gamma_i$) are
identical to the corresponding quantities of the true FMDP $M$, and hence to the current
approximate FMDP $\hat M_t$; (2) for all $a\in A$, $i\in\{1,\ldots,m\}$ and $\bx[\Gamma_i^a] \in
\b X[\Gamma_i^a]$, for any $y_i\in X_i$, the corresponding transition probability component is
$P^K_{t,i} (y_i | \bx[\Gamma_i],a) :=$
\[
   \left\{%
    \begin{array}{ll}
        \hat P_{t,i} (y_i | \bx[\Gamma_i],a), & \quad\hbox{if $(\bx[\Gamma_i],a)\in K_t$;} \\
        P_i (y_i | \bx[\Gamma_i],a), & \quad\hbox{if $(\bx[\Gamma_i],a)\not\in K_t$.} \\
    \end{array}%
    \right.
\]

Note that FMDPs $M^{K_t}$ and $\hat M_t$ are very close to each other: unknown state-action
components have identical transition functions by definition, while for known components, $\sum_{y
\in X_i} \left| P^K_{t,i}(y|\bx[\Gamma_i], a) - \hat P_{t,i}(y| \bx[\Gamma_i], a) \right| \leq
\frac{\epsilon(1-\gamma)}{m\gamma V_0}$. Consequently, for all $(\bx,a)$,
\begin{equation} \label{es:simlem_condition}
 \sum_{\by \in \b X} \left| P^K_{t}(\by|\bx, a) - \hat P_{t}(\by| \bx, a) \right| \leq
 \frac{\epsilon(1-\gamma)}{\gamma V_0} \,.
\end{equation}
For an arbitrary policy $\pi$, let $\b v^{K}_\pi$ and $\hat{\b v}_\pi$ be the value functions (the
fixed points of the approximate Bellman equations) of $\pi$ in $M^{K_t}$ and $\hat M_t$,
respectively. By a suitable variant of the Simulation Lemma (see supplementary material) that
works with the approximate Bellman equations, we get that whenever (\ref{es:simlem_condition})
holds, $\maxnorm{\b v^{K}_\pi-\hat{\b v}_\pi} \leq \epsilon$.

\subsubsection{The FOIM model is optimistic}

First of all, note that FOIM is not directly using the empirical transition probabilities $\hat
P_{t,i}$, but it is more optimistic; it  gives some chance for
getting to the garden of Eden state $x_E$: $\hat P^\textit{FOIM}_{t,i} (y_i | \b x[\Gamma_i],a) = $
\[
 \left\{%
\begin{array}{ll}
    \frac{k_{t\!,i}}{k_{t\!,i}+1} \hat{P}_{t,i}
 (y_i | \b x[\Gamma_i],a) , & \hbox{if $y_i \neq x_E$;} \\
    \frac{1}{k_{t\!,i}+1}, & \hbox{otherwise,} \\
\end{array}%
\right.
\]
where we introduced the shorthand $k_{t,i}=k_t(\bx[\Gamma_i],a)$.

Now, we show that
\begin{align} \nonumber
 & Q^\times(\b x,a) -
\Bigl[ R (\b x,a) + \gamma \sum_{\b y\in X} \hat{P}^\textit{FOIM}_t(\b y | \b x,a)  V^\times(\b
y) \Bigr] \\ \label{es:Qtimes_inequality}
 &\leq \Theta(\epsilon(1-\gamma)) \,,
\end{align}
or equivalently,
\begin{align*}
  &\sum_{\b y\in X}  (P(\b y \mid \b x,a) - P^\textit{FOIM}_t(\b y \mid \b x,a))V^\times(\by)
  \\
  &\geq
  -\sum_{i=1}^{m} {\textstyle \max (\frac{\beta}{\sqrt{k_{t,i}}}, \frac{\epsilon(1-\gamma)}{m} )\cdot\frac{k_{t,i}+1}{k_{t,i}} V_0 + \frac{1}{k_{t,i}} V_E.}
\end{align*}
Every term in the right-hand side is larger than $-\epsilon(1-\gamma)/m$, provided that we can
prove the slightly stronger inequality
\[ \textstyle
  - \max (\frac{\beta}{\sqrt{k_{t,i}}}, \frac{\epsilon(1-\gamma)}{m} )\cdot2 V_0 + \frac{1}{k_{t,i}} V_E
  \geq -\frac{\epsilon(1-\gamma)}{m} \,.
\]
First note that if the second term dominates the max expression, then the inequality
is automatically true, so we only have to deal with the situation when the first term dominates. In
this case, the inequality takes the form
$
  - \frac{\beta}{\sqrt{k_{t,i}}}\cdot2 V_0 + \frac{1}{k_{t,i}} V_E
  \geq -\frac{\epsilon(1-\gamma)}{m} \,,
$
which always holds because of our choice of $R_E$.

We show by induction that $V^{(t)}(\b x) \geq V^\times(\b x)-\Theta(\epsilon)$ and
$Q^{(t)}(\b x,a) \geq Q^\times(\b x,a)-\Theta(\epsilon)$ for all $t=0,1,2,\ldots$ and all $(\bx,a) \in
\b X\times A$. The inequalities hold for $t=0$. When moving from step $t$ to $t+1$,
\begin{align*}
&Q^{(t+1)} (\bx,a) =
  R (\b x,a) + \gamma \sum_{\b y\in \b X} \hat{P}_t(\b y \mid \b x,a)  V^{(t)}(\b y) \\
  &\geq
  R (\b x,a) + \gamma \sum_{\b y\in \b X} \hat{P}_t(\b y \mid \b x,a)  (V^\times(\b y)-\Theta(\epsilon)) \\
  &\geq
  Q^\times(\b x,a)-\gamma\Theta(\epsilon)-\Theta((1-\gamma)\epsilon)
\end{align*}
for all $(\bx,a)$, where we applied the induction assumption and eq.~(\ref{es:Qtimes_inequality}).
Consequently,
$
\max_{a\in A} Q^{(t+1)} (\bx,a) \geq \max_{a\in A} Q^\times (\bx,a) - \Theta(\epsilon)
$
for all $\bx$. Note that according to our assumptions, all entries of $H$ are nonnegative as well
as the entries of $G=\mathcal{N}(H^T)$, so multiplication by rows of $HG$ is a monotonous
operator, furthermore, all rows sum to 1, yielding
\begin{align*}
  & \sum_{\bx\in \b X} [HG]_{\by,\bx} \max_{a\in A} Q^{(t+1)} (\bx,a) \\
  & \geq \sum_{\bx\in \b X} [HG]_{\by,\bx} (\max_{a\in A} Q^\times (\bx,a) - \Theta(\epsilon)),
\end{align*}
that is, $V^{(t+1)}(\b x) \geq V^\times(\b x)-\Theta(\epsilon)$.

\subsubsection{Proximity of value functions}

The rest of the proof is standard, so we give here a very rough sketch only. We define a cutoff
horizon $H := \Theta(\frac{R_E}{1-\gamma} \log \frac{1}{\epsilon(1-\gamma)})$ and an escape
event $A$ which happens at timestep $t$ if the agent encounters an unknown transition in the next $H$ steps. We
will separate two cases depending on whether $\Pr(A)$ is smaller than $\frac{\epsilon(1-\gamma)}{R_E}$ or
not. If the probability of escape is low, then we can show that $Q^\textit{FOIM}(\bx_t,a_t) \geq
Q^\times(\bx_t,a_t)-\Theta(\epsilon)$. Otherwise, if $\Pr(A)$ is large, then an unknown
state-action component is found with significant probability. However, this can happen only at most $mN_f\asize
k_0$ times (because all components become known after $k_0$ visits), which is polynomial, so the second
case can happen only a polynomial number of times.

Finally, we remind that the statements are true only with probability
$1-\Theta(\delta)$. To round off the proof, we note that we are free to choose the constant in
the definition of $R_E$ (as it is hidden in the $\Theta(\cdot)$ notation), so we set it in a way
that $\Theta(\epsilon)$ and $\Theta(\delta)$ become at most $\epsilon$ and $\delta$, respectively.

%
%
%

\section{Discussion}

 FOIM is conceptually very simple: the exploration-exploitation dilemma is resolved
\emph{without any} explicit exploration, action selection is always greedy. The model update
and model solution are also at least as simple as the
alternatives found in the literature. Further, FOIM has some favorable theoretical properties. FOIM is
the first example to an RL algorithm that has a polynomial per-step computational complexity in
FMDPs. To achieve this, we had to relax the near-optimality of the FMDP planner.
The particular planner we used, FVI, runs in polynomial time, it does reach a bounded error, and the looseness of the bound depends on
the quality of basis functions. In almost all time steps, FOIM gets $\epsilon$-close to the FVI
value function with high probability (for any pre-specified $\epsilon$). The number of timesteps
when this does not happen is polynomial.\footnote{Note that in general there may be some
hard-to-reach states that are visited after a very long time only, so not all steps will be
near-optimal after a polynomial number of steps. This issue was analyzed by
\citet{Kakade03Sample}, who defined an analogue of ``probably approximately correctness'' for
MDPs.}

From a practical point of view, calling an FMDP model-solver in each iteration
could be prohibitive.  However, the model and the value function usually change very
little after a single model update, so we may initialize FVI with the previous value function, and
a few iterations might be sufficient.

\small
\section*{Acknowledgements}
This work has been supported by the  EC
NEST 'Perceptual Consciousness: Explication and Testing'
grant under contract 043261. Opinions and errors in this
manuscript are the author's responsibility, they do not
necessarily reflect the opinions of the EC or other project
members. The first author has been partially supported by the Fulbright Scholarship.

%
%
%
%

%

%
%


\appendix

\section{The Proof of Theorem 5.1}

\subsection{General lemmas}

\begin{lem} (Azuma's Inequality) If the random variables $X_1, X_2, \ldots$ form a
martingale difference sequence, meaning that $E[X_k|X_1,X_2, \ldots,X_{k-1}] = 0$ for all $k$, and
$|X_k| \leq b$ for each $k$, then
\[
  \Pr \left[ \sum_{i=1}^k X_i \geq a \right] \leq \exp \left( -
  \frac{a^2}{2b^2k}\right)
\]
and
\[
  \Pr \left[ \left|\sum_{i=1}^k X_i\right| \geq a \right] \leq 2\exp \left( -
  \frac{a^2}{2b^2k}\right)
\]
\end{lem}

\begin{lem}[Theorem 3 of \citet{Strehl07Model-Based}] \label{lem:visitcounts_strehl}
Fix a probability factor $i$ and a pair $(\bx[\Gamma_i],a) \in \b X[\Gamma_i]\times A$. Let $\hat
P_i(\cdot| \bx[\Gamma_i^a], a)$ be the empirical distribution of $P_i(\cdot| \bx[\Gamma_i], a)$
after $k_i$ visits to $(\bx[\Gamma_i],a)$. Then for all $\epsilon_1>0$, the $L_1$-error of the
approximation will be small with high probability:
\[
  \Pr\left( \sum_{y \in X_i} \left| P_i(y|\bx[\Gamma_i], a) - \hat P_i(y| \bx[\Gamma_i], a) \right| > \epsilon_1 \right)
  \leq 2^n \exp\left(-\frac{k_i\epsilon_1^2}{2}\right)
\]
\end{lem}

\begin{cor} \label{lem:visitcounts_to_Pi_L1}

For any $\delta_1>0$, define
\begin{equation} \label{e:beta_def}
  \beta(\delta_1) :=  \sqrt{2(\log \frac{1}{\delta_1}+ n\log 2)} \,.
\end{equation}
Then with probability at least $1-\delta_1$,
\[
   \sum_{y \in X_i} \left| \hat P_i(y|\bx[\Gamma_i], a) - P_i(y| \bx[\Gamma_i], a) \right| \leq \frac{\beta(\delta_1)}{\sqrt{k_i}}.
\]
\end{cor}

\begin{lem} \label{lem:visitcounts_to_Pi_L1_alltime}
Let $\delta_2 >0$, $\epsilon_2 >0$. Let
\[
   \delta' := C'\delta_2\epsilon_2^2/{\log \frac{1}{\delta_2\epsilon_2}}  \,
\]
with some suitable constant $C'$. Then
\[
  \Pr\left( \sum_{y \in X_i} \left| P_i(y|\bx[\Gamma_i], a) - \hat P_{t,i}(y| \bx[\Gamma_i], a) \right| > \max(\frac{\beta(\delta')}{\sqrt{k_t}}, \epsilon_2) \textrm{ for any $t=1,2,\ldots$}
  \right)\leq \delta \,,
\]
that is, the probability is very low that the approximate transition probabilities \emph{ever} get
very far from their exact values.
\end{lem}
 \bproof
By the union bound, the above probability is at most
\begin{eqnarray} \label{e:epsilondelta_probsum}
   \sum_{k=1}^\infty \Pr\left( \sum_{y \in X_i} \left| P_i(y|\bx[\Gamma_i], a) - \hat P_{t,i}(y| \bx[\Gamma_i], a) \right| > \max(\frac{\beta(\delta')}{\sqrt{k_t}}, \epsilon_2)
  \right) \,.
\end{eqnarray}
We will cut the sum into two parts, the cutting point $k_0$ is a constant to be determined later.
Define the auxiliary constants
\begin{eqnarray*}
  e &:=& \frac{1}{1-\exp\bigl(\frac{-\epsilon_2^2}{2}\bigr)} \\
  \delta' &:=& \frac{\delta_2}{k_0+e}\\
\end{eqnarray*}
Let $k_0$ such that $\frac{\beta(\delta')}{\sqrt{k_i}}$ becomes smaller than $\epsilon_2$ after
$k_0$ terms, that is,
 \(
  \frac{\beta(\delta')}{\sqrt{k_0}} \leq \epsilon_2,
 \)
 or equivalently,
\begin{eqnarray*}
  k_0 &\geq& \frac{\beta^2(\delta')}{\epsilon_2^2}
  = \frac{1}{\epsilon_2^2} \left( 2 \log \frac{1}{\delta'} + n \log 2 \right) \\
  &=& \frac{1}{\epsilon_2^2} \left( 2 \log (k_0+e) + 2 \log \frac{1}{\delta_2} + n \log 2 \right)
  \,.
\end{eqnarray*}
Using the very loose inequality $\log x \leq cx -1 - \log c$ with $c=\frac{\epsilon_2^2}{4}$, we
get that the above inequality holds if the stronger inequality
\begin{eqnarray*}
  k_0 &\geq& \frac{1}{2} (k_0+e) +\frac{4}{\epsilon_2^2} + \frac{4}{\epsilon_2^2}\log\frac{4}{\epsilon_2^2}+ \frac{1}{\epsilon_2^2} \left( 2 \log \frac{1}{\delta_2} + n \log 2 \right)
  \,,
\end{eqnarray*}
holds, that is, for
 \(
  k_0 \geq \Theta\left(\frac{1}{\epsilon_2^2}  \log \frac{1}{\delta_2\epsilon_2}\right) \,.
 \)
While this is a lower bound on $k_0$, this also means that there is a constant $C$ such that
\begin{equation} \label{e:k_0}
  k_0 = C \frac{1}{\epsilon_2^2}  \log \frac{1}{\delta_2\epsilon_2}
\end{equation}
satisfies the inequality.

Using the above facts and Corollary~\ref{lem:visitcounts_to_Pi_L1}, the sum of terms up to $k_0$ is
bounded by
\begin{eqnarray*}
  && \sum_{k=1}^{k_0-1} \Pr\left( \sum_{y \in X_i} \left| P_i(y|\bx[\Gamma_i], a) - \hat P_{k,i}(y| \bx[\Gamma_i], a) \right| > \frac{\beta(\delta')}{\sqrt{k_i}}
  \right) \\
  && \leq \sum_{k_i=1}^{k_0-1} \delta'  = (k_0-1)\delta'
   \leq \frac{k_0}{k_0+e}\delta_2 \\
\end{eqnarray*}

For the second part, note that the error probability for $k_0$ visits is at most $2^n
\exp\left(-\frac{k_0\epsilon_2^2}{2}\right) \leq \delta'$ by the definition of $k_0$. Therefore,
by Lemma~\ref{lem:visitcounts_strehl}, the sum of terms above $k_0$ is at most
\begin{eqnarray*}
  &&  \sum_{k_i=k_0}^\infty 2^n
  \exp\left(-\frac{k_i\epsilon_2^2}{2}\right)
   =  2^n
  \exp\left(-\frac{k_0\epsilon_2^2}{2}\right) \sum_{k_i=k_0}^\infty
  \exp\left(-\frac{(k_i-k_0)\epsilon_2^2}{2}\right) \\
  &&  \leq \delta' \sum_{k'=0}^\infty
  \exp\left(-\frac{k'\epsilon_2^2}{2}\right)
  \leq \delta' \frac{1}{1-\exp\left(-\frac{\epsilon_2^2}{2}\right)}
  = \delta'e = \delta_2 \frac{e}{k_0+e} \,.
\end{eqnarray*}
Consequently, the full sum is at most $\frac{k_0}{k_0+e}\delta_2 +\frac{e}{k_0+e} \delta_2 =
\delta_2$.

To complete the proof, note that $e = \Theta\left( \frac{1}{\epsilon_2^2}\right)$ (which follows
easily from the fact that $1+x \leq \exp(x) \leq 1+2x$ for $x\in[0,1]$). Furthermore, recall that
$k_0 = \Theta\left(\frac{1}{\epsilon_2^2}  \log \frac{1}{\delta_2\epsilon_2}\right)$, so $\delta'
= \delta_2 / \Theta\left(\frac{1}{\epsilon_2^2}  \log \frac{1}{\delta_2\epsilon_2}\right)$, that
is, $\delta' = \Theta\left(\delta_2\epsilon_2^2/{\log \frac{1}{\delta_2\epsilon_2}} \right)$, as
required.
 \eproof

The following lemma is almost identical to Corollary 1 of \citep{Strehl07Model-Based}, the only
change is that we allow different $\epsilon$s and $\delta$s for different components. The original
proof of \citet{Strehl07Model-Based} carries through with this modification in an unchanged
manner, so it is omitted here.
\begin{lem}\label{lem:Pi_to_P_L1}
Fix a pair $(\bx,a)\in \b X \times A$. Suppose that all probability factors are approximated well
in $L_1$-norm, i.e., for all $i$, there exist $\epsilon_{3,i}>0, \delta_{3,i}\geq 0$ such that
\[
   \sum_{y_i \in X_i} \left|\hat P_i(y_i | \bx[\Gamma_i],a) - P_i(y_i | \bx[\Gamma_i],a)\right| \leq \epsilon_{3,i}.
\]
with probability at least $1-\delta_{3,i}$. Then
\[
   \sum_{\b y \in \b X} \left|\hat P(\b y | \bx, a) - P(\b y | \bx, a) \right| \leq \sum_{i=1}^{m} \epsilon_{3,i}
\]
with probability at least $1-\sum_{i=1}^{m} \delta_{3,i}$.
\end{lem}

%

The previous lemma bounds the error for a single state. The following corollary extends the
results, showing that the probability of a large approximation error \emph{anywhere} in the state
space is low.
\begin{lem} \label{lem:Pi_to_P_L1_allstates}
Let $\delta_4 >0$, $\epsilon_4 >0$. Let
\[
   \delta' := C'\frac{\delta_4\epsilon_4^2}{m^3N_f\asize}/{\log \frac{m^2N_f\asize}{\delta_4\epsilon_4}}  \,
\]
with some suitably defined constant $C'$. Then with probability at least $1-\delta_4$,
\[
   \sum_{\b y \in \b X} \left|\hat P_t(\b y | \bx, a) - P(\b y | \bx, a) \right| \leq
   \sum_{i=1}^m \max (\frac{\beta(\delta')}{\sqrt{k_t(\bx[\Gamma_i],a)}}, \frac{\epsilon_4}{m})
\]
for any $t=1,2,\ldots$ and any $(\bx,a)\in \b X\times A$.

\end{lem}
 \bproof
Fix a component $(\bx[\Gamma_i],a)\in \b X[\Gamma_i] \times A$. By applying
Lemma~\ref{lem:visitcounts_to_Pi_L1_alltime} to this component with $\epsilon_2 =
\frac{\epsilon_4}{m}$ and $\delta_2 = \frac{\delta_4}{mN_f\asize}$, we get that
\begin{equation} \label{e:P_i_global_prob}
  \sum_{y \in X_i} \left| P_i(y|\bx[\Gamma_i], a) - \hat P_{t,i}(y| \bx[\Gamma_i], a) \right| \leq \max(\frac{\beta(\delta')}{\sqrt{k_t(\bx[\Gamma_i],a)}}, \frac{\epsilon_4}{m})
\end{equation}
for all $t$ with probability at least $1-\frac{\delta_4}{mN_f\asize}$. There are $mN_f\asize$ different
components, so the probability that (\ref{e:P_i_global_prob}) is ever violated for any of them is
still less than $\delta_4$. If no components violate (\ref{e:P_i_global_prob}), then we can apply
Lemma~\ref{lem:Pi_to_P_L1} to all $(\bx,a)\in \b X\times A$ and all timesteps $t=1,2,\ldots$,
proving the statement of the lemma.
 \eproof

\begin{defn}[known state-action components] For any $\epsilon_4>0, \delta_4>0$, let
\[
    KB(\epsilon_4,\delta_4) := C \frac{m^2}{\epsilon_4^2}  \log \frac{m^2N_f\asize}{\delta_4\epsilon_4} \,,
\]
where $C$ is a suitable constant defined by eq.~(\ref{e:k_0}).
The pair
$(\bx[\Gamma_i],a) \in \b X[\Gamma_i]\times A$ is \emph{$(\epsilon_4,\delta_4)$-known}, if
\[
    k(\bx[\Gamma_i],a) \geq KB(\epsilon_4,\delta_4) \,.
\]
\end{defn}

Note that $KB(\epsilon_4,\delta_4)$ is the number of visit counts from which on the second term
quantity in the maximum expressions dominates the first one. Therefore, after more than
$KB(\epsilon_4,\delta_4)$ visits to a component, we can really feel confident that it is known: if
all components were known, then all the approximate transition probabilities of the FMDP would be
within $\epsilon_4$ $L_1$-distance from the true values, with less than $\delta_4$ total
probability of an error.

%
%

\subsection{Some bounds for value functions}

Let us assume that all rewards fall between 0 and $\rmax$. In that case, the maximum possible
value of a state is $\vmax := \frac{\rmax}{1-\gamma}$.

\begin{lem}
Consider an FMDP with transition functions $\{{P}^a\}$, and let $\pi$ be an arbitrary policy. Let
$\tilde{\b v}^\pi$ the fixed point of iteration
\[
  \tilde{\b v}^\pi = H G \sum_{a\in A} \pi(\cdot,a)  \bigl( \b r^a + \gamma P ^a
   \tilde{\b v}^\pi \bigr).
\]
There exists a universal bound
\[
  V_0 := \frac{(3-\gamma)}{(1-\gamma)} \vmax = O\left(\frac{\rmax}{(1-\gamma)^2}\right)
\]
for which $\maxnorm{\tilde{\b v}^\pi} \leq V_0$.
\end{lem}
 \bproof
Let $\b v^\pi$ the solution of the exact Bellman-equations: ${\b v}^\pi = \sum_{a\in A}
\pi(\cdot,a)  \bigl( \b r^a + \gamma P ^a
   {\b v}^\pi \bigr).$
Then
\begin{eqnarray*}
  \maxnorm{\tilde{\b v}^\pi - {\b v}^\pi}
 &=& \maxnorm{(HG)\sum_{a\in A} \pi(\cdot,a)  \bigl( \b r^a + \gamma P ^a
   \tilde{\b v}^\pi \bigr) - \sum_{a\in A} \pi(\cdot,a)  \bigl( \b r^a + \gamma P ^a
   \b v^\pi \bigr)} \\
 &\leq& \maxnorm{(HG)\sum_{a\in A} \pi(\cdot,a)  \bigl( \b r^a + \gamma P ^a
   \tilde{\b v}^\pi \bigr) - (HG)\sum_{a\in A} \pi(\cdot,a)  \bigl( \b r^a + \gamma P ^a
   \b v^\pi \bigr)} \\
 &&+ \maxnorm{(HG)\sum_{a\in A} \pi(\cdot,a)  \bigl( \b r^a + \gamma P ^a
   {\b v}^\pi \bigr) - \sum_{a\in A} \pi(\cdot,a)  \bigl( \b r^a + \gamma P ^a
   \b v^\pi \bigr)} \\
 &\leq& \gamma\maxnorm{HG}\maxnorm{\tilde{\b v}^\pi - {\b v}^\pi} + \maxnorm{(HG)\b v^\pi - \b
 v^\pi} \\
 &\leq& \gamma\maxnorm{\tilde{\b v}^\pi - {\b v}^\pi} + \maxnorm{HG}\maxnorm{\b v^\pi} + \maxnorm{ \b
 v^\pi} \,,
\end{eqnarray*}
so $\maxnorm{\tilde{\b v}^\pi - {\b v}^\pi} \leq \frac{2}{1-\gamma}\vmax$. Therefore,
\begin{eqnarray*}
 \maxnorm{\tilde{\b v^\pi}} &\leq& \maxnorm{\tilde{\b v}^\pi - \b v^\pi} + \maxnorm{\b v^\pi} \\
 &\leq& (\frac{2}{1-\gamma} +1) \vmax = \frac{(3-\gamma)}{(1-\gamma)} \vmax.
\end{eqnarray*}

%
%
 \eproof

\begin{lem}
The AVI-optimal value function $\b v^\times$ is also bounded by $V_0$:
\[
  \maxnorm{\b v^\times} \leq V_0 \,.
\]
\end{lem}
 \bproof
For any $G$ satisfying $\maxnorm{HG}\leq 1$, 
\begin{eqnarray*}
 \maxnorm{\b v^\times} &\leq& \maxnorm{\b v^\times - \b v^*} + \maxnorm{\b v^*} \leq
 \frac{1}{1-\gamma} \maxnorm{HG\b v^* - \b v^*} + \maxnorm{\b v^*} \\
 &\leq& \frac{1}{1-\gamma} \maxnorm{HG}\maxnorm{\b v^*} + \frac{1}{1-\gamma}\maxnorm{\b v^*} + \maxnorm{\b
 v^*}\\
 &\leq& (\frac{2}{1-\gamma} + 1) \maxnorm{\b v^*} \leq  (\frac{2}{1-\gamma} + 1) \vmax =
 V_0,
\end{eqnarray*}
where we used the triangle-inequality and eq.~(5) of the
main paper in the first line.
 \eproof

\subsection{The known-state FMDP}

\begin{lem}[Simulation lemma with function approximation] \label{lem:simlem_fapp}
Let $\epsilon_5>0$. Consider two FMDPs $M$ and $\hat M$ with joint transition probabilities $P$
and $\hat P$, otherwise identical. For any policy $\pi:\b X\times A$, consider the corresponding
value functions $\b v_\pi$ and $\hat{\b v}_\pi$. If
\[
  \sum_{\by\in \b X}\left| \hat P(\b y | \bx, a) - P(\b y | \bx, a) \right| \leq \frac{\epsilon_5(1-\gamma)}{\gamma
  V_0}
\]
for all $(\bx,a)\in \b X\times A$, then
\[
  \maxnorm{ \hat{\b v}_\pi - \b v_\pi } \leq \epsilon_5.
\]
\end{lem}
 \bproof
Let $\Delta := \maxnorm{ \b v_\pi - \hat{\b v}_\pi }$, and let $(\bx_\Delta,a_\Delta)$ be a
state-action pair for which $\max_{a\in A} \maxnorm{P^a \b v_\pi - \hat P^a \hat{\b v}_\pi}$ takes
its maximum, i.e.,
\[
 \max_{a\in A} \maxnorm{P^a \b v_\pi - \hat P^a \hat{\b v}_\pi} =
 \left| \sum_{\by\in \b X}  P(\by|\bx_\Delta,a_\Delta) V_\pi(\by)
         -  \hat P(\by|\bx_\Delta,a_\Delta) \hat V_\pi(\by)\right|.
\]
Using this,
\begin{eqnarray*}
  \Delta &=& \maxnorm{
     H G \sum_{a\in A} \pi(\cdot,a)  \bigl( \b r^a + \gamma P^a \b v_\pi \bigr)
     - H G \sum_{a\in A} \pi(\cdot,a)  \bigl( \b r^a + \gamma \hat P^a \hat{\b v}_\pi \bigr)} \\
  &\leq& \maxnorm{HG} \gamma \max_{a\in A} \maxnorm{P^a \b v_\pi - \hat P^a \hat{\b v}_\pi} \\
  &\leq& \gamma \max_{a\in A} \maxnorm{P^a \b v_\pi - \hat P^a \hat{\b v}_\pi} \\
  &=& \gamma \left| \sum_{\by\in \b X}  P(\by|\bx_\Delta,a_\Delta) V_\pi(\by)
         -  \hat P(\by|\bx_\Delta,a_\Delta) \hat V_\pi(\by)\right| \\
  &\leq& \gamma \left| \sum_{\by\in \b X}  \bigl[P(\by|\bx_\Delta,a_\Delta) - \hat P(\by|\bx_\Delta,a_\Delta)\bigr] V_\pi(\by) \right|
         +\gamma \left| \sum_{\by\in \b X} \hat P(\by|\bx_\Delta,a_\Delta) \bigl[V_\pi(\by)- \hat
         V_\pi(\by)\bigr]\right| \\
  &\leq& \gamma  \sum_{\by\in \b X}  \Bigl| P(\by|\bx_\Delta,a_\Delta) - \hat P(\by|\bx_\Delta,a_\Delta)\Bigr| V_0
         +\gamma  \sum_{\by\in \b X} \hat P(\by|\bx_\Delta,a_\Delta) \maxnorm{ \b v_\pi - \hat{\b v}_\pi} \\
  &\leq& \frac{\epsilon_5(1-\gamma)}{\gamma V_0} \gamma V_0
         +\gamma \Delta = (1-\gamma)\epsilon_5 + \gamma \Delta.
\end{eqnarray*}

 \eproof



\begin{defn}[known-state FMDP]
Let $\epsilon>0, \delta>0$ be arbitrary probabilities. Consider an FMDP $M$ and a series of FMDPs
$\hat M_t$ with joint transition probabilities $P$ and $\hat P_t$, otherwise identical.
Furthermore, let $K_t$ be the set of $(\epsilon,\delta)$-known $(\bx[\Gamma_i^a],a)$ pairs. Define
the FMDP $M^{K_t}$ so that
\begin{itemize}
 \item its state and action space, rewards, and the decompositions of the transition probabilities
(i.e., the dependency sets $\Gamma_i$) are identical to $M$ and $\hat M_t$,
 \item for all $a\in A$, $i\in\{1,\ldots,m\}$ and $\bx[\Gamma_i^a] \in \b X[\Gamma_i^a]$, for any $y_i\in
X_i$, the corresponding transition probability component is
\[
  P^K_{t,i} (y_i | \bx[\Gamma_i],a) := \left\{%
    \begin{array}{ll}
        \hat P_{t,i} (y_i | \bx[\Gamma_i],a), & \quad\hbox{if $(\bx[\Gamma_i],a)\in K_t$ (known pairs);} \\
        P_i (y_i | \bx[\Gamma_i],a), & \quad\hbox{if $(\bx[\Gamma_i],a)\not\in K_t$ (unknown pairs).} \\
    \end{array}%
    \right.
\]
\end{itemize}
\end{defn}



%

\begin{lem} \label{lem:betaineq}
Let $\epsilon_6>0$, $\delta_6>0$. Suppose that FOIM is executed on an FMDP $M= (\b
X,A,P,R,\gamma,\{\Gamma_i\},\{Z_j\})$. If the initial value of the garden of Eden state is at
least
\begin{equation} \label{e:R_E_condition}
  R_E \geq
c\cdot \frac{m\rmax^2}{(1-\gamma)^3\epsilon_6} \left[\log \frac{mN_f \asize}{\epsilon_6\delta_6}\right]
\end{equation}
with some constant $c$, then with probability at least $1 - \delta_6$,
\begin{equation} \label{e:Qtimes_inequality}
 Q^\times(\b x,a) -
\left[ R (\b x,a) + \gamma \sum_{\b y\in X} \hat{P}^\textit{FOIM}_t(\b y \mid \b x,a)  V^\times(\b
y) \right]
 \leq \epsilon_6
\end{equation}
for all $(\bx,a) \in \b X\times A$ and all $t=1,2,\ldots$
\end{lem}
 \bproof
By Lemma~\ref{lem:Pi_to_P_L1_allstates} (with setting $\epsilon_4=\epsilon_6$ and $\delta_4 =
\delta_6$), for any $t=1,2,\ldots$ and any $(\bx,a)\in \b X\times A$,
\[
   \sum_{\b y \in \b X} \left|\hat P_t(\b y | \bx, a) - P(\b y | \bx, a) \right| \leq
   \sum_{i=1}^m \max (\frac{\beta(\delta')}{\sqrt{k_t(\bx[\Gamma_i],a)}}, \frac{\epsilon_6}{m})
\]
with probability at least $1-\delta_6$, where
\begin{equation} \label{e:delta'_in_betaineqlemma}
 \delta' := C'\frac{\delta_6\epsilon_6^2}{m^3N_f \asize}/{\log \frac{m^2N_f \asize}{\delta_6\epsilon_6}} \,.
\end{equation}

Fix a state-action pair $(\b x,a)$, and let us use the shorthand $k_{t,i} = k_t(\bx[\Gamma_i],a)$
and $\beta= \beta(\delta')$. Define the transition probabilities $\hat P^\textit{FOIM}_{t,i}$ as
the empirical approximate probabilities with the hypothetical visit to the garden of Eden state.
Note that
\[
 \hat P^\textit{FOIM}_{t,i} (y_i \mid \b x[\Gamma_i],a) = \left\{%
\begin{array}{ll}
    \frac{k_{t,i}}{k_{t,i}+1} \hat{P}_{t,i}
 (y_i \mid \b x[\Gamma_i],a) , & \hbox{if $y_i \neq x_E$;} \\
    \frac{1}{k_{t,i}+1}, & \hbox{otherwise,} \\
\end{array}%
\right.
\]
so
\[
  \sum_{\b y\in X}  (P(\b y \mid \b x,a) - P^\textit{FOIM}_t(\b y \mid \b x,a))V^\times(\by) \geq
  -\sum_{i=1}^{m} \max (\frac{\beta}{\sqrt{k_{t,i}}}, \frac{\epsilon_6}{m} )\cdot\frac{k_{t,i}+1}{k_{t,i}} V_0 + \frac{1}{k_{t,i}} V_E.
\]
We will prove that every term in the right-hand side is larger than $-\epsilon_6/m$. We are going
to prove the slightly stronger inequality
\[
  - \max (\frac{\beta}{\sqrt{k_{t,i}}}, \frac{\epsilon_6}{m} )\cdot2 V_0 + \frac{1}{k_{t,i}} V_E
  \geq -\frac{\epsilon_6}{m} \,.
\]
First of all, note that if the second term dominates the max expression, then the inequality is
automatically true, so we only have to deal with the situation when the first term dominates.
In this case, the inequality to
prove becomes
\[
  - \frac{\beta}{\sqrt{k_{t,i}}}\cdot2 V_0 + \frac{1}{k_{t,i}} V_E
  \geq -\frac{\epsilon_6}{m} \,.
\]

After multiplication by $k_{t,i}$ and taking the derivative, we get that that the left-hand side
takes its minimum where
\[
 -\beta  V_0 \frac{1}{\sqrt{k_{t,i}}} + \frac{\epsilon_6}{m} = 0,
\]
that is, for
\[
 k_{t,i} = \left( \frac{m\beta V_0}{\epsilon_6} \right)^2.
\]
Substituting the minimum place to the inequality, we get that it always holds if
\[
  - \frac{\beta\epsilon_6}{m\beta V_0}\cdot2 V_0 + \frac{\epsilon_6^2}{(m\beta V_0)^2} V_E
  \geq -\frac{\epsilon_6}{m},
\]
that is, if
\begin{eqnarray}
  V_E &\geq& \frac{m}{\epsilon_6} ( V_0)^2 \beta^2 \label{e:V_E_derivation}\\
  &=& \frac{m}{\epsilon_6} ( V_0)^2 (2\log \frac{m^3N_f \asize\log\frac{m^2N_f \asize}{\delta_6\epsilon_6}}{C'\delta_6\epsilon_6^2}+ n\log 2)
    \nonumber \\
 &=& \Theta\left( \frac{m\rmax^2}{(1-\gamma)^4\epsilon_6} \left[\log \frac{m^3N_f \asize\log\frac{m^2N_f \asize}{\delta_6\epsilon_6}}{\epsilon_6^2\delta_6}\right]
  \right)  \nonumber \\
 &=& \Theta\left( \frac{m\rmax^2}{(1-\gamma)^4\epsilon_6} \left[\log \frac{mN_f \asize}{\epsilon_6\delta_6}\right]
  \right) \,, \nonumber
\end{eqnarray}
which holds by the assumption of the lemma. During the transformations, we used the definition of
$\beta = \beta(\delta')$ in eq.~(\ref{e:beta_def}), the definition of $\delta'$ in
eq.~(\ref{e:delta'_in_betaineqlemma}), the fact that $ V_0 =
\Theta\left(\frac{\rmax}{(1-\gamma)^2}\right)$ and that $n$ is a small constant hidden by the
$\Theta(\cdot)$ notation. By noting that $V_E = R_E/(1-\gamma)$, the proof of the lemma is
complete.



%
\eproof

The following result shows that FOIM preserves the optimism of the value function with high
probability.

\begin{lem} \label{lem:optimismpreserved}
Let $\epsilon_7>0$, $\delta_7>0$. Suppose that FOIM is used with $\epsilon_6 =
(1-\gamma)\epsilon_7$, $\delta_6=\delta_7$ and $R_E$ satisfying (\ref{e:R_E_condition}). Then,
with probability at least $1 - \delta_7$, $V^{(t)}(\b x) \geq V^\times(\b x)-\epsilon_7$ and
$Q^{(t)}(\b x,a) \geq Q^\times(\b x,a)-\epsilon_7$ for all $t=0,1,2,\ldots$ and all $(\bx,a) \in
\b X\times A$.
\end{lem}

According to Lemma~\ref{lem:betaineq}, eq.~(\ref{e:Qtimes_inequality}) holds for all $(\bx,a) \in
\b X\times A$ and all $t$ with high probability. So, except for an error event (with probability
at most $\delta_7$), We can proceed with the following induction on the number of DP-updates.
Initially, $V^{(0)}(\bx) \geq V^\times(\bx) - \epsilon_7$. When moving from step $t$ to $t+1$,
\begin{eqnarray*}
Q^{(t+1)} (\bx,a) &=&
  R (\b x,a) + \gamma \sum_{\b y\in \b X} \hat{P}_t(\b y \mid \b x,a)  V^{(t)}(\b y) \\
  &\geq&
  R (\b x,a) + \gamma \sum_{\b y\in \b X} \hat{P}_t(\b y \mid \b x,a)  (V^\times(\b y)-\epsilon_7) \\
  &\geq&
  Q^\times(\b x,a)-\gamma\epsilon_7-(1-\gamma)\epsilon_7
\end{eqnarray*}
for all $(\bx,a)$, where we applied the induction assumption and lemma~\ref{lem:betaineq}.
Consequently,
\[
\max_{a\in A} Q^{(t+1)} (\bx,a) \geq \max_{a\in A} Q^\times (\bx,a) - \epsilon_7
\]
for all $\bx$. Note that according to our assumptions, all entries of $H$ are nonnegative as well
as the entries of $G=\mathcal{N}(H^T)$, so multiplication by rows of $HG$ is a monotonous
operator, furthermore, all rows sum to 1, yielding
\[
  \sum_{\bx\in \b X} [HG]_{\by,\bx} \max_{a\in A} Q^{(t+1)} (\bx,a) \geq \sum_{\bx\in \b X} [HG]_{\by,\bx} (\max_{a\in A} Q^\times (\bx,a) -
  \epsilon_7),
\]
that is, $V^{(t+1)}(\b x) \geq V^\times(\b x)-\epsilon_7$ with prob. $1-\delta_7$.
  \eproof

%

\subsection{Proximity of value functions}

In the following, we will show that whenever the algorithm remains in the known region of the
FMDP, its value function is very close to the AVI-optimal $Q^\times$. The two value functions will
be related to each other through a sequence of other value functions.

%


Let us fix $\epsilon>0$, $\delta>0$, and let $\epsilon_8 := \epsilon/4$, $\delta_8 := \delta/2$. Let
\[
  H := \frac{R_E}{1-\gamma} \log \frac{1}{\epsilon_8(1-\gamma)}
\]
be the $\epsilon_8$-horizon time. For a given point during the execution of the algorithm, let $A$
denote the event that the algorithm will encounter an unknown transition in the next $H$ steps. We
will separate two cases depending on whether $\Pr(A)$ is small or large. Firstly, assume that
\[
  \Pr(A) < \frac{\epsilon_8(1-\gamma)}{R_E}
\]

Let $M$ denote the true (and unknown) FMDP, and fix a pair $(\bx_1,a_1)\in \b X \times A$. Let
$Q_M^\textit{FOIM}(\bx_1,a_1)$ be the expected reward collected by FOIM in $M$, and let
$Q_M^\textit{FOIM}(\bx_1,a_1,H)$ be the $H$-step truncated version.

\textbf{Statement 1.} \emph{$Q_M^\textit{FOIM}(\bx_1,a_1) \geq Q_M^\textit{FOIM}(\bx_1,a_1,H)$. }

 \bproof
Because of our assumption that all rewards are nonnegative, truncation removes only nonnegative
terms.
 \eproof

Let $M^{K_t}$ be the known-state FMDP defined by the $(\epsilon_8,\delta_8)$-known components.

\textbf{Statement 2.} \emph{$Q_M^\textit{FOIM}(\bx_1,a_1,H) \geq
Q_{M^{K_t}}^\textit{FOIM}(\bx_1,a_1,H) - \epsilon_8$. }

 \bproof
On known states, $M$ and $M^{K_t}$ are identical (by the definition of $M^{K_t}$), so the
collected rewards are identical, too. If the algorithm encounters an unknown state-action pair,
the difference of the two value functions may be as large as $R_E/(1-\gamma)$. However, the
probability that an unknown pair is found in the next $H$ steps is at most
$\frac{\epsilon_8(1-\gamma)}{R_E}$ by assumption, so
\[
  Q_M^\textit{FOIM}(\bx_1,a_1,H) \geq Q_{M^{K_t}}^\textit{FOIM}(\bx_1,a_1,H) -
\Pr(A) \frac{R_E}{1-\gamma} \geq Q_{M^{K_t}}^\textit{FOIM}(\bx_1,a_1,H) - \epsilon_8 \,.
\]
 \eproof

\textbf{Statement 3.} \emph{$Q_{M^{K_t}}^\textit{FOIM}(\bx_1,a_1,H) \geq
Q_{M^{K_t}}^\textit{FOIM}(\bx_1,a_1) - \epsilon_8$. }

 \bproof
This is a simple restatement of the fact that $H$ is an $\epsilon_8$-horizon time.
 \eproof

%

Let $\hat{M}_t$ be the approximate FMDP built by FOIM, and suppose that
\begin{equation} \label{e:R_E_condition_2}
  R_E =
  c\cdot \frac{m\rmax^2}{(1-\gamma)^4\epsilon_8} \left[\log \frac{mN_f \asize}{(1-\gamma)\epsilon_8\delta_8}\right] \,.
\end{equation}

\textbf{Statement 4.} \emph{ $Q_{M^{K_t}}^\textit{FOIM}(\bx_1,a_1) \geq
Q_{\hat{M}_t}^\textit{FOIM}(\bx_1,a_1) - \epsilon_8$. }

 \bproof
Follows from Lemma~\ref{lem:simlem_fapp} by substituting .
 \eproof

\textbf{Statement 5.} \emph{If $R_E$ satisfies eq.~(\ref{e:R_E_condition_2}), then
$Q_{\hat{M}}^\textit{FOIM}(\bx_1,a_1) \geq Q^\times(\bx_1,a_1) - \epsilon_8$ with probability at
least $1-\delta_8$.}

 \bproof
This is basically the statement of Lemma~\ref{lem:optimismpreserved} with the assignment
$\epsilon_7 := \epsilon_8$ and $\delta_7 := \delta_8$.
 \eproof

%
%
%
%
%

Summing up statements 1--5, we get that for $\Pr(A)<\frac{\epsilon_8(1-\gamma)}{R_E}$,
\begin{equation} \label{e:FOIM_nearoptimality}
    Q_M^\textit{FOIM}(\bx_1,a_1) \geq Q^\times(\bx_1,a_1) - 4\epsilon_8 = Q^\times(\bx_1,a_1) - \epsilon
\end{equation}
with probability $1-\delta_8$.

\subsection{Finding unknown regions}

We will now show that whenever
\begin{equation} \label{e:Pr_A}
\Pr(A)>\frac{\epsilon_8(1-\gamma)}{R_E} \,,
\end{equation}
we will make a significant update to the model with relatively high probability. We will use the
following simple consequence of the Hoeffding inequality [Thomas Walsh, personal communication]:
\begin{lem} \label{lem:coinflip}
Suppose a weighted coin, when flipped, has probability $p
> 0$ of landing with heads up. Then, for any positive integer $k$ and real number $\delta \in (0,
1)$, there exists a number $m = O(\frac{k}{p} \log\frac{1}{\delta})$, such that after $m$ tosses,
with probability at least $1 - \delta$, we will observe $k$ or more heads.
\end{lem}

\begin{lem} \label{lem:Pr_A_occurences}
With probability $1-\delta$, (\ref{e:Pr_A}) will occur at most
\[
  O\left( \frac{   \rmax^2 m^4 N_f \asize   }{\epsilon^4(1-\gamma)^4} \log^3\frac{1}{\delta} \log^2 \frac{mN_f \asize}{\epsilon}
  \right)
\]
times.
\end{lem}

 \bproof
Let $N_A$ be the number of timesteps when (\ref{e:Pr_A}) holds, and for all $n=1,\ldots,N_A$, let
\[
  a_{t_n} := \left\{%
\begin{array}{ll}
    1, & \hbox{if an unknown pair $(\bx_{t_n}[\Gamma_i],a) \not \in K$ was encountered;} \\
    0, & \hbox{otherwise.} \\
\end{array}%
\right.
\]
The number of model updates is simply $\sum_{n=1}^{N_A} a_{t_n}$, which is at most $m N_f \asize \cdot
KB(\epsilon_8,\delta_8) = O(m N_f \asize \cdot KB(\epsilon,\delta))$. On the other hand, by
Lemma~\ref{lem:coinflip}, $\sum_{n=1}^{N_A} a_{t_n}$ will be at least $m N_f \asize \cdot
KB(\epsilon_8,\delta_8)$ with probability at least $1-\delta$ after
\begin{eqnarray*}
 O\left( \frac{m N_f \asize \cdot KB(\epsilon/4,\delta)}{\Pr(A)} \log\frac{1}{\delta} \right)
 &=&  O\left( \frac{R_E m N_f \asize \cdot KB(\epsilon,\delta)}{\epsilon(1-\gamma)} \log\frac{1}{\delta}
  \right) \\
 &=&  O\left( \frac{R_E m^3 N_f \asize   }{\epsilon^3(1-\gamma)} \log\frac{1}{\delta} \log \frac{m^2N_f \asize}{\delta\epsilon}
  \right) \\
 &=&  O\left( \frac{   \rmax^2 m^4 N_f \asize   }{\epsilon^4(1-\gamma)^4} \log^3\frac{1}{\delta} \log^2 \frac{mN_f \asize}{\epsilon}
  \right) \\
\end{eqnarray*}
steps.
 \eproof

Putting the two cases together, we get the following:
\begin{thm}
Let $\epsilon>0$ and $\delta>0$, and suppose FOIM is initialized with
\[
R_E = 4c\cdot \frac{m\rmax^2}{(1-\gamma)^4\epsilon} \left[\log \frac{8mN_f \asize}{(1-\gamma)\epsilon\delta}\right]
\]
 With probability at least
$1-\delta$, the number of timesteps when FOIM makes non-AVI-near-optimal moves, i.e., when
\[
  Q^\textit{FOIM}(\bx_t,a_t) < Q^\times(\bx_t,a_t) - \epsilon \,,
\]
is bounded by
\[
  O\left( \frac{   \rmax^2 m^4 N_f \asize   }{\epsilon^4(1-\gamma)^4} \log^3\frac{1}{\delta} \log^2 \frac{mN_f \asize}{\epsilon}
  \right) \,.
\]
\end{thm}
 \bproof
By eq.~(\ref{e:FOIM_nearoptimality}), with probability $1-\delta/2$, FOIM makes AVI-near-optimal
moves whenever $\Pr(A)$ is small. The number of times this does not happen is bounded by
Lemma~\ref{lem:Pr_A_occurences} with probability $1-\delta/2$, and is exactly the bound given by the
statement of the theorem.
 \eproof

\bibliographystyle{plain}

\end{document}